\documentclass[letterpaper, 10 pt, conference]{ieeeconf}  % Comment this line out if you need a4paper

\IEEEoverridecommandlockouts                              % This command is only needed if 
\usepackage{cite}

\usepackage{amsmath,amssymb,amsfonts}

\usepackage{gensymb}

\usepackage{graphicx}

\usepackage{textcomp}

\usepackage{xcolor}

\usepackage{verbatim}

\usepackage{afterpage}

\usepackage{cite}

\usepackage{overpic}

\usepackage{psfrag}

\usepackage{latexsym}

\usepackage{bm}

\usepackage{cases}

\usepackage{array}

\usepackage{fancyhdr}

\usepackage{setspace}

\usepackage{subfigure}

\usepackage{url}

\usepackage{multirow}

\usepackage{epstopdf}

\usepackage{epsfig}

\usepackage{fancybox}

\usepackage{textcomp}

\usepackage[ruled,linesnumbered]{algorithm2e}

\def\BibTeX{{\rm B\kern-.05em{\sc i\kern-.025em b}\kern-.08em
    T\kern-.1667em\lower.7ex\hbox{E}\kern-.125emX}}

\DeclareMathOperator*{\argminA}{arg\,min} % Jan Hlavacek
\newcommand{\x}[0]{\mathbf{x}}
\newcommand{\T}[0]{\mathbf{T}}
\newcommand{\f}[0]{\mathbf{f}}
\newcommand{\C}[0]{\mathbf{C}}
\newcommand{\s}[0]{\mathrm{sim}}
\begin{document}

\title{Efficient WiFi LiDAR SLAM for Autonomous Robots in Large Environments}%SLAM based on radio fingerprints and scan-matching laser scans loop closures\\}

\author{Khairuldanial Ismail, Ran Liu, Zhenghong Qin, Achala Athukorala, Billy Pik Lik Lau, Muhammad Shalihan,\\
Chau Yuen, and U-Xuan Tan
\thanks{K. Ismail, R. Liu, A. Athukorala, B. P. L. Lau, M. Shalihan, C. Yuen, and U-X. Tan are with the Engineering Product Development Pillar, Singapore University of Technology and Design, 8 Somapah Rd, Singapore, 487372.
 {\{\tt\small ran\_liu, achala\_chathuranga, billy\_lau, yuenchau, uxuan\_tan\}@sutd.edu.sg}.}
\thanks{Z. Qin is with the School of Information Engineering, Southwest University of Science and Technology, Mianyang, China, 621010.}
\thanks{This work is partially supported by the National Key R\&D Program of China 2019YFB1310805 and the National Science Foundation of China 12175187.}
}

\maketitle
\begin{abstract}
Autonomous robots operating in indoor and GPS denied environments can use LiDAR for SLAM instead. However, LiDARs do not perform well in geometrically-degraded environments, due to the challenge of loop closure detection and computational load to perform scan matching. Existing WiFi infrastructure can be exploited for localization and mapping with low hardware and computational cost. Yet, accurate pose estimation using WiFi is challenging as different signal values can be measured at the same location due to the unpredictability of signal propagation. Therefore, we introduce the use of WiFi fingerprint sequence for pose estimation (i.e. loop closure) in SLAM. This approach exploits the spatial coherence of location fingerprints obtained while a mobile robot is moving. This has better capability of correcting odometry drift. The method also incorporates LiDAR scans and thus, improving computational efficiency for large and geometrically-degraded environments while maintaining the accuracy of LiDAR SLAM. We conducted experiments in an indoor environment to illustrate the effectiveness of the method. The results are evaluated based on Root Mean Square Error (RMSE) and it has achieved an accuracy of 0.88m for the test environment. 
\end{abstract}
%\begin{IEEEkeywords}
%SLAM, localization, WiFi, LiDAR, loop closure
%\end{IEEEkeywords}
\section{Introduction}
\begin{comment}
This enables high level tasks that are vital for autonomous functions such as navigation and path planning \cite{intro_slam1}\cite{intro_slam2}\cite{intro_slam3}\cite{intro_slam4}. However, prior knowledge of the environment is required to perform localization. This is obtained with mapping which describes the landmarks or features within an environment to a robot. Maps of the environment are commonly obtained beforehand, however, this is not possible under certain circumstances such as disaster areas or large-scale environments which pose as a challenge for robots to perform indoor positioning \cite{intro_slam5}\cite{intro_slam6}.  
\end{comment}
Simultaneous Localization and Mapping (SLAM) enables autonomous robots to localize in unknown environments by simultaneously obtaining its pose and generating a map of its environment. Research have been extensively investigating and developing different SLAM techniques to achieve high performance and efficiency. SLAM methods that provide accurate pose estimates are usually based on visual cameras and Light Detection and Ranging (LiDAR). However, these methods do not work well in large and geometrically-degraded environments due to the difficulty of robust loop closure detection. Furthermore, these methods are vulnerable to dynamic environments and occlusions.  

In outdoor environments, accurate positioning is made possible with the availability of Global Positioning System (GPS). However, this method of localization is not available for indoor environment due to multipath propagation and no clear line-of-sight between GPS receivers and GPS satellites \cite{gps}. Research done in the past decade suggests alternatives to GPS by proposing methods that utilize Ultra-Wideband (UWB)\cite{intro_slam6}, Bluetooth Low Energy (BLE)\cite{ble}, Long-Term Evolution (LTE)\cite{lte}, and WiFi\cite{wifi}. The wide deployment of WiFi in indoor environment allows for accessibility of the signal without requiring additional infrastructure. Hence, it becomes the signal of choice for the proposed method.  

\begin{figure}[h]
    \centering
    \includegraphics[width = 3.4in]{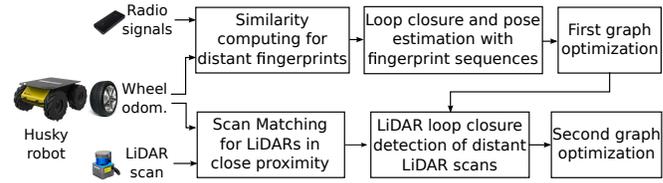}
    \caption{Overview of the proposed WiFi LiDAR SLAM.}
    \label{fig:overview}
\end{figure}

This paper proposes a method which involves a two level optimization process for SLAM. An overview of the proposed approach is illustrated in Figure \ref{fig:overview}. The first level uses graph optimization on loop closures based on pose estimation using fingerprint sequences. It does not require prior knowledge on distribution of the access points (APs) or a labor-intensive data collection phase unlike fingerprinting-based method. The second level further improves the accuracy of the trajectory obtained. It uses graph optimization on loop closures based on scan matching of LiDAR scans. Accurate pose estimations can also be obtained using LiDAR scans and scan matching. While this is computationally intensive, the trajectory produced in the first step reduces the number of scans required for scan matching to those that are in close proximity. Thus, improving computational efficiency while preserving accuracy. Furthermore, the first level provides robustness against geometrically-degraded environments where LiDAR-based SLAM performs less reliably. We conducted experiments in an indoor carpark environment and showed that the proposed method has managed to achieve accurate pose estimation of a Clearpath Husky UGV. The contribution of this paper includes: 1) proposing a WiFi SLAM method using loop closures based on WiFi fingerprint sequences; 2) proposing an additional step of improving WiFi SLAM using LiDAR; and 3) conducting experiments to illustrate the performance, efficiency, and robustness of the proposed method. 

We organise the paper as follows. In Section II, we introduce related works in SLAM and localization techniques that use WiFi or LiDAR. In Section III, the paper explains the methodology of WiFi SLAM using loop closures based on WiFi fingerprint sequences and the additional step to incorporate LiDAR. In Section IV, the paper describes the experiment conducted and evaluates the results. 
Section V concludes this paper with a discussion of the future work.

\section{Related works}
WiFi signals present an opportunity to be exploited for indoor positioning purposes. However, these signals were not designed for estimating pose and thus techniques have been investigated and studied to enable this feature. These techniques can be broadly categorised into fingerprinting-based, model-based, and machine learning-based approaches. Fingerprinting-based techniques \cite{fprint1} rely on forming the fingerprint map which consists of WiFi data mapped to spatial coordinates that represents physical locations. While this method can be accurate, the process of creating the fingerprint map can be very tedious especially over large areas, which is not sustainable in the long run \cite{fprint2}. 
Authors in \cite{wifi_enhenced_mapping1} proposed to fuse WiFi fingerprints into visual SLAM algorithm. To improve runtime efficiency, 
visual matching is performed when a similar WiFi cluster is found. 
Authors in \cite{wifi_enhenced_mapping2} used WiFi to determine the coarse orientation between different sub-maps during distributed online mapping.

Another technique involves obtaining a model that can represent the characteristics of the signal in the environment. The work presented in \cite{propmodel}, uses received signal strength (RSS)-based localization with device free localization techniques based on their modified log-distance model which employs a maximum likelihood technique to estimate position. However, parameters such as the transmitting power and path loss exponent are practically unknown. It is almost impossible to obtain them due to the multiple path propagation phenomenon in indoor environments. WiFi SLAM in \cite{gplvm1}, mapped high-dimensional signal strength into two-dimensional latent space with a Gaussian process. Authors in \cite{gplvm2} improved this by utilizing graph-based SLAM. Both methods assume the signal strength at two close locations to be similar and the measurement likelihood can be modeled using a Gaussian process. In contrast, the proposed method does not require any model to represent signal propagation or signal strength distribution. 

Machine learning-based techniques rely on training a model using prior data to estimate position. The method described in \cite{ml} utilizes random forests algorithms for WiFi localization and extends the method by using Gaussian process latent variable model (GPLVM) to represent the signal strength distribution. As mentioned, the proposed method requires neither prior knowledge of the environment nor a model representation of the signal strength distribution. 
%other techniques for indoor positioning methods that utilizes WiFi includes computing time of arrival \cite{toa}, time difference of arrival \cite{tdoa}, and angle of arrival \cite{aoa}. Despite good accuracy, implementing these techniques is complex and requires special equipment. 

LiDARs are able to generate laser scans which measure relative distances to its surrounding environment \cite{lidar1} and represent the structural maps of environments, which are suitable for SLAM. Methods that are widely used for LiDAR-based SLAM are GMapping \cite{lidar2}, HectorSLAM\cite{lidar3}, and Cartographer\cite{lidar4}. These methods, however, are not suitable for large areas \cite{lidar5}. The scan matching procedures can be computationally intensive to perform in larger areas. Geometrically-degraded environments also pose a challenge for scan matching to identify a proper match for loop closures. By incorporating WiFi measurements into the conventional LiDAR SLAM, our proposed method is shown to be more robust in large-scale and geometrically-degraded environments. 

\section{Methodology}
This paper proposes a novel approach of using WiFi and, subsequently, LiDAR for SLAM in an unknown environment. This section explains the methodology which involves: 1) an overview of the SLAM framework for this method; 2) fingerprint similarity computation; 3) detecting loop closure candidates based on fingerprint sequences; and 4) integrating LiDAR into WiFi SLAM. 

\subsection{Overview of graph-based SLAM framework}
We adopt a graph-based approach which performs SLAM through a maximum likelihood estimation using pose graphs. The node $\mathbf{x}_t = (x_t,y_t,\theta_t)$ in the graph is defined as the pose of the robot at time $t$. We use $\mathbf{x} = \{\mathbf{x}_1,...,\mathbf{x}_T\}$ to denote the path of the robot up to time $T$. 
$\mathbf{f}_t$ represents WiFi measurement recorded at location $\mathbf{x}_t$ at time $t$. 
$\mathbf{f}_t$ consists of RSS values from $L$ access points, i.e., $\mathbf{f}_t=({f}_{t,i},..., {f}_{t,L})$. 
%LiDAR laser scans are recorded concurrently for the method explained later in Section II.E. 
The graph uses sequential odometry measurements as odometry-based constraints and loop closures as observation-based constraints. 
In particular, the constraint between two nodes $i$ and $j$ is defined as a rigid body transformation with mean $z_{ij}$ and covariance $\Sigma_{ij}$.
%, which are obtained from either odometry or sensor measurements.
Loop closure represents a situation when the robot has revisited a previously observed location. It is determined by aligning sensor observations between two non-consecutive poses. As odometry is susceptible to drifting, loop closure corrects the accumulated odometric error. The proposed method recognises loop closure based on WiFi fingerprint sequences and incorporates these constraints into graph-based SLAM. Scan matching is used to infer the additional LiDAR constraints to improve the localization accuracy. We perform graph optimization by minimizing Equation \ref{eq:graph_eqn} in order to obtain accurate trajectory as the best configuration of $\mathbf{x}$, given the constraints.
\begin{equation} \label{eq:graph_eqn}
    \sum_{i,j\in C}(\mathbf{z}_{ij} - \mathbf{\hat{z}}_{ij}(\mathbf{x}_i,\mathbf{x}_j))^T\Sigma_{ij}^{-1}(\mathbf{z}_{ij} - \mathbf{\hat{z}}_{ij}(\mathbf{x}_i,\mathbf{x}_j)),
\end{equation}
where $C$ denotes the set containing all pairs of constraints. 

\subsection{Fingerprint similarity}
Radio fingerprint uses a set of radio signals from different base stations to represent a location.
Similar to appearance-based approach, the radio fingerprint is assumed to be unique in an environment, 
which is widely used for localization in the research community and industrial applications.
In contrast to visual-based localization, where heavy computation is involved for feature extraction and matching, 
the time required for comparing fingerprint similarity is negligible, 
as unique MAC (Media Access Control) address reported by the AP can be served as ubiquitous feature for the localization.

Given two fingerprints $\mathbf{f}_i$ and $\mathbf{f}_j$, we use $H$ to represent the number of common APs 
(i.e., the APs appear in both fingerprints). In addition, let $L_i$ and $L_j$ denote total number of APs in $\mathbf{f}_i$ and $\mathbf{f}_j$, respectively. 
The similarity function $sim(\mathbf{f}_i, \mathbf{f}_j)$ gives a positive value between 0 and 1, 
that considers the detection likelihood and the signal strength likelihood:
\begin{equation} \label{eq:meanmodel}
\s(\mathbf{f}_i, \mathbf{f}_j)= \underbrace{\frac{H}{L_i+L_j-H}}_{\text{Detection likelihood}} \cdot \underbrace{\frac{1}{H} \prod_{n=1}^{H}  \exp(-\frac{(f_{i,n}-f_{j,n})^2}{2\sigma^2})}_{\text{Signal strength likelihood}},
\end{equation}
where $\sigma^2$ represents the received signal variance, which is set to 36dBm according to the findings in \cite{gps}.

\begin{figure}[h]
    \centering
    \includegraphics[width = 0.9\linewidth]{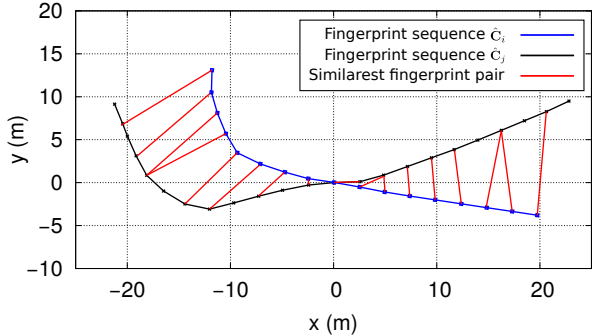}
    \caption{Illustration of two fingerprint sequences and the nearest fingerprint pairs found with $k=1$.}
    \label{fig:fingerprint_sequence}
\end{figure}

\subsection{Loop closure and pose estimation with fingerprint sequence}
\label{sec:sequence_sim}
The accuracy of fingerprint-based approach is often affected by the unpredictability of radio propagation.
To address this issue, we present an approach for pose estimation 
based on the matching of two fingerprint sequences by exploiting their spatial coherence.
A slide window $w$ is used to segment the track into fingerprint sequences. 
In particular, we define a sequence $\C_t$ at time $t$ as a collection of sequential poses with a 
window size of $w$, i.e., $\C_t=\{\x_{t'}\}$, where $t-\frac{w}{2} \le t' \le t+\frac{w}{2}$. 
Since our goal is to estimate the relative pose between two fingerprint $\x_i$ and $\x_j$, 
therefore we use the relative pose of $\x_{t'}$ with respect to $\x_{t}$ in a sequence $\C_t$, i.e., 
$ \hat{\C}_t=\{\x_{t}^{-1}\x_{t'}\}$.
We first check the similarity between two fingerprints $\f_i$ and $\f_j$. 
If the similarity is higher than a predefined threshold (for example 0.3), 
we use our sequence-based approach to determine the relative pose between $\x_i$ and $\x_j$. 
This is achieved by finding the best configuration of the pose (${\T}^*$, including 2D displacement and rotation) 
through minimizing the residual error of two fingerprint sequences:
\begin{equation} \label{eq:pose_estimation}
{\T}^*=\argminA_{\T} \frac{1}{w} \sum_{\tau=-\frac{w}{2}}^{\frac{w}{2}} \text{dist}(\T(\x_i^{-1}\x_{i+\tau}), \x_{j}^{-1}\x_{j^*}),
\end{equation}
where the function $\text{dist}(\cdot)$ computes the Euclidean distance between two locations. 
$\x_{j^*}$ represents the estimated position in sequence $\hat{\C_{j}}$ for a given fingerprint $\f_{i+\tau}$ in sequence $\hat{\C_{i}}$.
Given a location fingerprint $\f_{i+\tau}$, 
we compute its similarities to those fingerprints located 
in a time window $2w$ with respect to $\f_j$. 
As a result, we obtain the $k$ reference fingerprints $\f_{\pi(1)},...,\f_{\pi(k)}$, whose similarities best match the fingerprint $\f_{i+\tau}$. We approximate $\x_{j^*}$ as the weighted mean among these $k$-nearest locations in similarity space: 
\begin{equation} \label{eq:xj}
\x_{j^*}=\frac{1}{\sum_{l=1}^{k}{\s(\f_{i+\tau},\f_{\pi(l)})} } \sum_{l=1}^{k}{\s(\f_{i+\tau},\f_{\pi(l)}) \cdot \x_{\pi(l)}},
\end{equation}
where $\pi(l)$ is within range $[j-\frac{w}{2}, j+\frac{w}{2}]$. An example of the fingerprint sequences and the similarest fingerprint pairs found with $k=1$ is shown in Figure\,\ref{fig:fingerprint_sequence}.
The minimization (i.e., Equation \ref{eq:pose_estimation}) is solved by singular value decomposition (SVD). 
We consider a valid loop closure $<\x_i, \x_j>$ if the average distance (i.e., dist($\cdot$) in Equation \ref{eq:pose_estimation}) for $\T^*$ is smaller than a threshold (3 meters). The covariance matrix of this constraint is set to be identity.

\subsection{Second graph optimization using LiDAR scans}
\label{sec:scan_matching}
Based on WiFi constraints inferred in Section \ref{sec:sequence_sim}, 
we perform the first pose graph optimization (i.e., WiFi SLAM) to obtain a coarse aligned trajectory of the robot. 
The Levenberg-Marquardt in g2o\footnote[1]{https://github.com/RainerKuemmerle/g2o} is used as solver to optimize the graph.
Afterwards, we perform the second pose graph optimization by minimizing Equation \ref{eq:graph_eqn} again based on the optimized trajectory from the first pose graph optimization with new constraints obtained from LiDAR scan matching.
The trajectory provided by the first graph optimization reduces the frequency to perform scan matching for loop closures while the second graph optimization improves its accuracy. A common used approach for scan matching is the iterative closest point (ICP) algorithm, which iteratively finds a transformation between two LiDAR scans to minimize the sum of square distance (i.e., fitness score) between the correspondence points. 
We consider the following two manners to incorporate LiDAR scans:
\begin{itemize}
\item \textbf{Scan matching of LiDARs in close proximity:} we perform ICP of the LiDAR scans in close proximity to correct the short term odometry errors. 
The odometry estimation is used as the initial guess for the ICP algorithm.
\item \textbf{LiDAR loop closure detection:} in addition to the evaluation of LiDAR scans in proximity, we perform the LiDAR scan matching for non-consecutive poses if the displacement of two poses is smaller than a threshold. 
This step determines loop closures between non-consecutive poses, which allows to correct the long term odometry error.
\end{itemize}
%We optimize Equation \ref{eq:graph_eqn} by the pose graph optimization algorithm. 
%The Levenberg-Marquardt in g2o is used as solver to optimize the graph.
The optimized trajectory is annotated with LiDAR scans to construct the grid map of the environment. The working flow of the proposed approach is described in Algorithm\,\ref{algo}.

\begin{algorithm}
\small
\label{algo}
\SetKw{KwWith}{with}
\SetKw{KwEach}{each}
\SetKw{KwOr}{or}
\SetKw{KwAnd}{and}
  \SetKwData{Left}{left}\SetKwData{This}{this}\SetKwData{Up}{up}

  \SetKwFunction{Union}{Union}\SetKwFunction{FindCompress}{FindCompress}

  \SetKwInOut{Input}{input}\SetKwInOut{Output}{output}

\KwData{Odometry $\{\x_t\}_{t=1}^{T}$, WiFi fingerprints $\{\f_t\}_{t=1}^{T}$, and LiDAR scans}

\KwResult{Optimized trajectory $\{\hat{\x}_t\}_{t=1}^{T}$}

\caption{The proposed approach for WiFi LiDAR SLAM}

\tcp{Loop closure and relative pose estimation based on WiFi fingerprint sequence (Sect.\,\ref{sec:sequence_sim})}

\For{$i\leftarrow 1$ \KwTo $T$}{
\tcp{Check accumulated distance and fingerprint similarity}
    \For{$j\leftarrow 1$ \KwTo $i$ \KwWith $\text{acc}(\x_i, \x_j) \ge 50$ \KwAnd $sim(\f_i, \f_j) \ge 0.3$}{    

   $\rhd$ Compute the relative pose ${\T}^{*}$ between $\x_i$ and $\x_j$ according to Equation \ref{eq:pose_estimation}\
 
   \If{Average distance computed for ${\T}^{*}$ is smaller than 3 meters }
   {
   $\rhd$ Add $<\x_i, \x_j>$ as loop closure
   }
    }

  }

$\rhd$ Obtain the optimized trajectory $\{\bar{\x}_t\}_{t=1}^{T}$ by graph optimization according to Equation \ref{eq:graph_eqn} \\

\tcp{LiDAR scan matching (Sect.\,\ref{sec:scan_matching}) for pose refinement}

\For{$i\leftarrow 1$ \KwTo $T$}{
\tcp{Check accumulated distance and relative distance of $\bar{\x}_i$ and $\bar{\x}_j$}
    \For{$j\leftarrow 1$ \KwTo $i$ \KwWith $\text{acc}(\x_i, \x_j) \le 1$ \KwOr $ ||\bar{\x}_i - \bar{\x}_j|| \le 5$}{    

   $\rhd$ LiDAR scan registration between $\x_i$ and $\x_j$ according to ICP\
 
   \If{Matching points $\ge$ half of the average number of points}
   {
   $\rhd$ Add $<\x_i, \x_j>$ as loop closure
   }
    }

  }

$\rhd$ Obtain the final trajectory $\{\hat{\x}_t\}_{t=1}^{T}$ by optimizing Equation \ref{eq:graph_eqn} \\

\end{algorithm}

\section{Experimental result}
\subsection{Setup}
\begin{figure}[h]
    \centering
    \includegraphics[width = 0.5\linewidth]{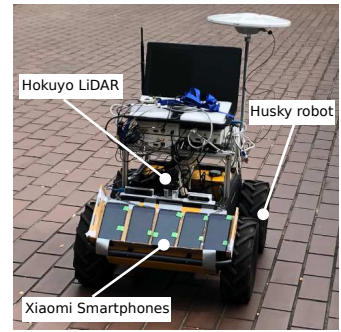}
    \caption{Experimental setup.}
    \label{fig:setup}
\end{figure}

The experiment was conducted at the Basement 2 carpark B (with a size of 30m$\times$200m) in Nanyang Technological University (NTU) campus. 
The experiment was performed using a Clearpath Husky UGV, as shown in Figure\,\ref{fig:setup}. 
A Hokuyo LiDAR UST-20LX was mounted on the robot to obtain laser scans. Xiaomi Max 3 smartphones were placed on the robot to scan WiFi access points at a frequency of 0.5Hz. Wheel odometry measurements were recorded at a frequency of 10Hz. The ground truth was obtained through Adaptive Monte Carlo Localization (AMCL) with a prior map, created using GMapping on data recorded over two hours in the environment. Throughout the experiment, the robot was driven at an average speed of 0.4m/s. The robot traveled over a distance of 1072.6m with a duration of about 2583 seconds. 
In total, 1046 MAC addresses are reported in the environment. 
In the next sections, we proceed to evaluate: 1) WiFi SLAM based on fingerprint sequence; 2) the effects on accuracy by integrating LiDAR scans; 3) the mapping performance as compared to GMapping; and 4) runtime.

\subsection{Evaluation of WiFi SLAM based on fingerprint sequence} % threshold described in section II.D
For this evaluation, we consider WiFi SLAM without integrating laser scans. 
Table \ref{tab:compare_window_size} evaluates the performance of relative pose estimation (including 2D position and orientation) using fingerprint sequence and WiFi SLAM by varying the window size $w$ from 20 to 120. 
We fix $k$=2 for this series of experiments. 
A large window size reduces the number of constraints and improves the relative pose estimation accuracy using fingerprint sequence, while computational time increases. 
However, it does not provide a better localization accuracy using WiFi SLAM, due to the reduced number of constraints.
At small window size, despite having more constraints and less computational time, it introduces higher uncertainties which does not improve or even deteriorates the performance of WiFi SLAM. 
Hence, we set a window size $w$ to 80 for the remaining evaluations. 
Table \ref{tab:compare_k} evaluates the accuracy and constraints of WiFi SLAM by varying the $k$ from 1 to 8. 
%We fix the widows size $w$=80 for this series of experiments. 
As seen from this table, $k=2$ achieves the best positioning accuracy. 
A too large or too small $k$ reduces the positioning accuracy. 

\begin{table}[]
\small
\caption{Evaluation of the proposed approach under different window size $w$.}
\label{tab:compare_window_size}
\begin{tabular}{|l|c|c|c|c|}
\hline
$w$ (window size)                                                                                             & 20   & 40   & 80   & 120   \\ \hline \hline
Number of loop closures                                                                          & 1038   & 620   & 396   & 257   \\ \hline
\begin{tabular}[c]{@{}l@{}}Position estimation accuracy\\with fingerprint sequence (m)\end{tabular} & 2.93   & 1.05   & 0.57   & 0.47   \\ \hline
\begin{tabular}[c]{@{}l@{}}Orientation estimation accuracy \\with fingerprint sequence (rads)\end{tabular} & 0.48   & 0.23   & 0.10   & 0.11   \\ \hline
Total computational time (s)                                                                            & 306   & 364   & 652   & 1138   \\ \hline
WiFi SLAM accuracy (m)                                                                          & 2.87 & 2.75 & 2.70 & 3.66 \\ \hline
\end{tabular}
\end{table}

\begin{table}[]
\small
\caption{Evaluation of the proposed approach under different settings of $k$.}
\label{tab:compare_k}
\begin{tabular}{|l|c|c|c|c|}
\hline
$k$ (reference fingerprints)                                                                                             & 1   & 2   & 4   & 8   \\ \hline \hline
Number of loop closure                                                                            & 426   & 396   & 320   & 265   \\ \hline
\begin{tabular}[c]{@{}l@{}}Position estimation accuracy\\with fingerprint sequence (m)\end{tabular} & 0.72   & 0.57   & 0.72   & 0.99   \\ \hline
\begin{tabular}[c]{@{}l@{}}Orientation estimation accuracy\\with fingerprint sequence (rads)\end{tabular} & 0.11   & 0.10   & 0.11   & 0.13   \\ \hline
WiFi SLAM RMSE (m)                                                                         & 2.83 & 2.70 & 2.77 & 3.83 \\ \hline
\end{tabular}
\end{table}

\begin{figure}[h]
    \centering
    \includegraphics[width = 0.6\linewidth]{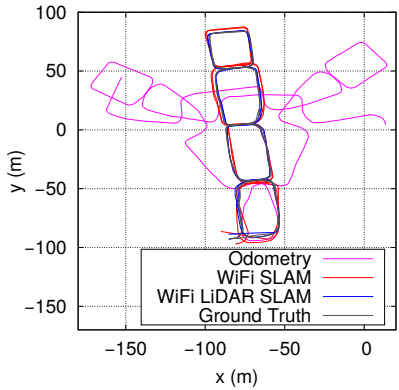}
    \caption{Visualization of ground truth, odometry, WiFi SLAM, and WiFi LiDAR SLAM.}
    \label{fig:track}
\end{figure}

\subsection{WiFi LiDAR SLAM by integrating LiDAR scans}
Integrating laser scans in proximity would improve the accuracy of wheel odometry. ICP-based scan matching of laser scans is enabled if the distance traveled by the robot is less than 1 meter. 
%When the distance threshold is too small, the error increases due to the less number of laser scans that are being integrated, while additional evaluation of laser scans increases the computational time.
Using the optimized trajectory obtained from the first pose graph optimization, we obtain loop closures by performing scan matching if relative distance between two poses is within a distance threshold, i.e., 5 meters. 
We discard a match if the number of correspondence points is less than half of the average number of points in two scans. 
Table \ref{tab:compare_gmapping} lists the positioning accuracy with different approaches. 
We compared our approach with the one \cite{binning} that uses a pre-trained similarity model for the WiFi loop closures. 
Our approach (with WiFi only) provides a localization accuracy of 2.7m, which is an improvement of 30.9\% compared to the similarity model-based approach (3.91m).
%Figure \ref{fig:percentage_scan} shows the accuracy, number of LiDAR constraints, and runtime with varying percentages of poses used.
Scan matching is time consuming to perform between all poses due to high sampling rate of the odometry. 
Hence, percentage of poses examined is limited to improve computational efficiency. 
With more poses evaluated, we obtained better accuracy. 
However, we observe that the accuracy does not continue to improve for evaluating more than 40\% of the poses. 
Whereas, time consumption increases linearly with the increase of the percentage of poses evaluated. 
Figure \ref{fig:track} shows the comparison of trajectories using different variations of the proposed methods with the ground truth.

\begin{figure}
\centering
\includegraphics[width=0.49\textwidth]{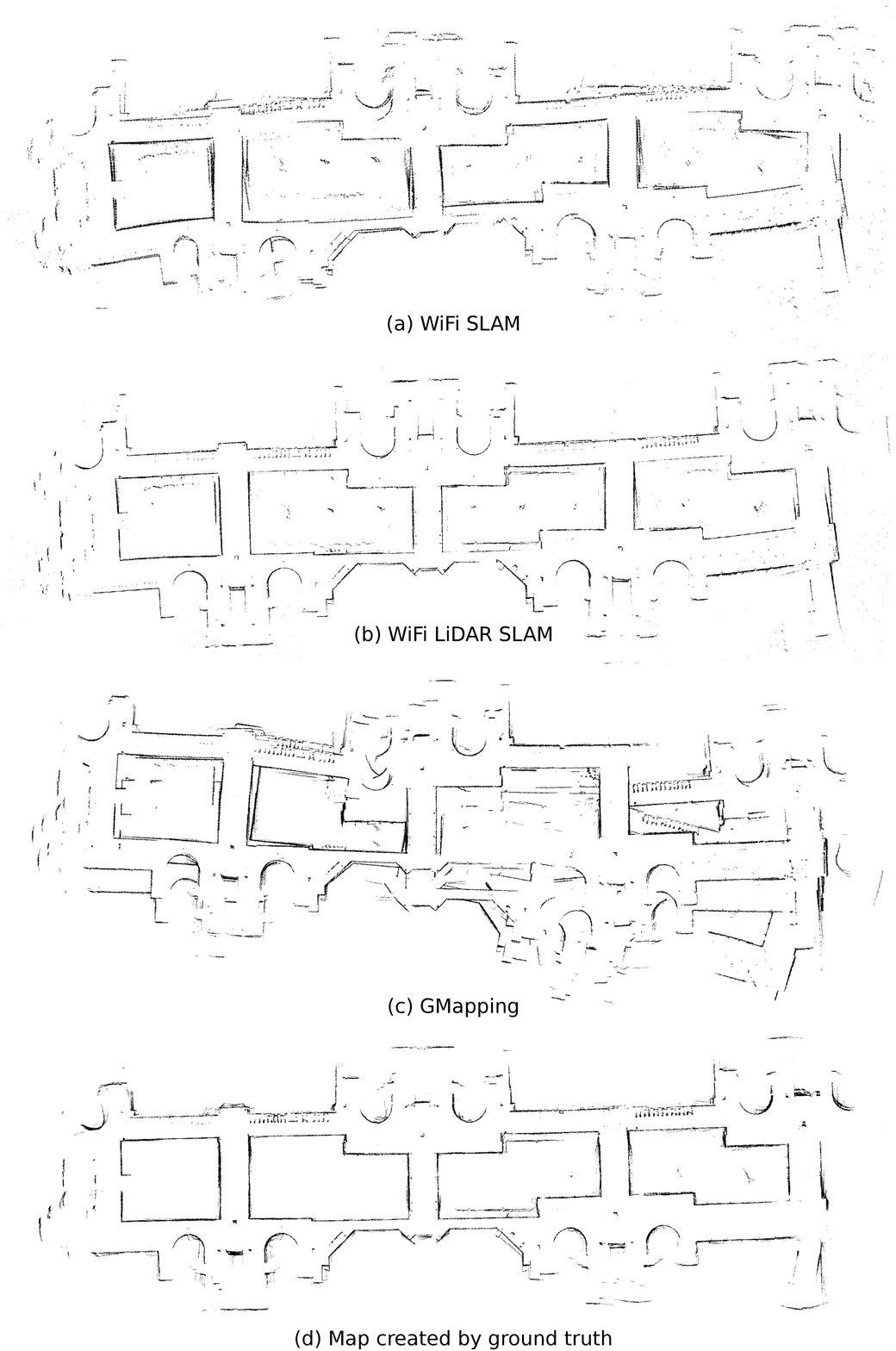}
\caption{
A comparison of the maps created by different approaches, i.e., WiFi SLAM, WiFi LiDAR SLAM, GMapping, and ground truth.
}
\label{fig:maps}
\end{figure}

\begin{table*}[]
\small
\centering
\caption{Comparison to GMapping and the mapping accuracy (meters) of our approach by using different sensory information.}
\label{tab:compare_gmapping}
\begin{tabular}{|c|c|c|c|c|c|c|}
\hline
Odometry & GMapping & \begin{tabular}[c]{@{}l@{}}Similarity\\model\cite{binning}\end{tabular} & WiFi & \begin{tabular}[c]{@{}l@{}}WiFi+\\close scans\end{tabular} & \begin{tabular}[c]{@{}l@{}}WiFi+close scans+\\10\% additional scans\end{tabular} & \begin{tabular}[c]{@{}l@{}}WiFi+close scans+\\40\% additional scans\end{tabular} \\ \hline
 54.7    &   4.31       &     3.91                                                               &  2.7    &          1.56                                                  &      0.92                                                     &       0.88                                                                          \\ \hline
\end{tabular}
\end{table*}
\begin{table}[]
\small
\caption{Evaluation of running time at different stages of our approach.}
\label{tab:running_time}
\begin{tabular}{|l|c|}
\hline
Stage                                                                                 & Time consumed in seconds \\ \hline
Data recording                                                                        & 2583.95                 \\ \hline
\begin{tabular}[c]{@{}l@{}}WiFi loop closure with\\ fingerprint sequence\end{tabular} & 364                        \\ \hline
First graph optimization                                                              & 2.76                        \\ \hline
\begin{tabular}[c]{@{}l@{}}Scan matching in close\\ proximity\end{tabular}            & 917.2                        \\ \hline
Additional 10\% laser scans                                                          & 3887.62                     \\ \hline
Second graph optimization                                                             & 2.84                        \\ \hline
\end{tabular}
\end{table}

\subsection{Evaluation of SLAM performance as compared to GMapping}
GMapping applies particle filtering for SLAM and is considered as one of the state-of-the-art SLAM algorithms. 
We refer to the work in \cite{gmapping_vs_carographer} for a comparison between GMapping and Cartographer, where the authors concluded that GMapping produced similar maps when compared to Cartographer.
We applied GMapping to the recorded dataset and compared with the different variations of the proposed method as shown in Table\,\ref{tab:compare_gmapping}. Our proposed method of SLAM involving the two level optimization produced a positioning accuracy of 0.88m by fusing the close LiDAR scans and 40\% of additional LiDAR scans for loop closure detection. GMapping failed to correct the large odometric errors in the large environments and hence did not give the most accurate results. 

Based on the optimization trajectory, we generate the occupancy grid maps by evaluating the respective grid using laser scans at optimized poses. 
We evaluated the map produced by the proposed method, GMapping, and the ground truth, as shown in Figure \ref{fig:maps}. 
The map by WiFi SLAM (see Figure \ref{fig:maps}(a)) is shown to be globally consistent but, with slight distortion in local areas. 
This can be seen from the non-straight boundaries representing the walls. The quality of the map is significantly improved by the incorporation of laser scan matching, which can be seen from Figure \ref{fig:maps}(b). The purpose of the map is to visually inspect quality of the optimized trajectory. It is important to mention that, to generate our ground truth map by GMapping, we have to design a cumbersome strategy for the robot to traverse different small regions and avoid large loops. This requires experienced users and a good familiarity with the environment, which might be difficult for general users especially in unknown environments.

\subsection{Evaluation of runtime}
We tested the proposed approach on an Intel Core i7-6700HQ CPU with 2.6GHz frequency and an RAM of 8GB. We set window size $w=80$ for the test. 
The percentage of poses evaluated for LiDAR loop closure detection is limited to 10\%. 
Optimization is performed after all loop closures are identified. Table \ref{tab:running_time} shows the computational time taken at each stage of the proposed method. From Table \ref{tab:running_time}, we observe that the time taken for WiFi loop closure detection and optimization are insignificant as compared to the data recording time. The time required for performing one WiFi similarity comparison of 44 MAC addresses is 0.029ms whereas, time required for one scan matching of 600 points is 13.04ms. Therefore, laser scan matching is the bottleneck of the proposed method. Applying state-of-the-art fast scan matching algorithms would lead to an improvement of the performance of the proposed method.

\section{Conclusion and discussion}
In this paper, we proposed an approach to efficiently and accurately perform SLAM for autonomous robots by the fusion of WiFi fingerprint sequence and LiDAR laser scans. Our approach produces a globally consistent trajectory based on WiFi measurements and then refines the trajectories by scan matching of laser scans. The performance of the system was verified in an indoor environment. 
The proposed method obtained an accuracy of 0.88m for the test environment by the fusion of WiFi and LiDAR. 
In the future, we would like to investigate the possibility of using the produced map to assist the navigation and path planning of an autonomous robot.


\begin{thebibliography}{99}

\begin{comment}


\bibitem{intro_slam1} A. Yassin et al., "Recent Advances in Indoor Localization: A Survey on Theoretical Approaches and Applications," in \textit{IEEE Communications Surveys \& Tutorials}, vol. 19, no. 2, pp. 1327-1346, Secondquarter 2017.

\bibitem{intro_slam2} S. He and S. -. G. Chan, "Wi-Fi Fingerprint-Based Indoor Positioning: Recent Advances and Comparisons," in \textit{IEEE Communications Surveys \& Tutorials}, vol. 18, no. 1, pp. 466-490, Firstquarter 2016.

\bibitem{intro_slam3} X. Guo, S. Zhu, L. Li, F. Hu and N. Ansari, "Accurate WiFi Localization by Unsupervised Fusion of Extended Candidate Location Set," in \textit{IEEE Internet of Things Journal}, vol. 6, no. 2, pp. 2476-2485, April 2019.

\bibitem{intro_slam4} W. Li, Z. Chen, X. Gao, W. Liu and J. Wang, "Multimodel Framework for Indoor Localization Under Mobile Edge Computing Environment," in \textit{IEEE Internet of Things Journal}, vol. 6, no. 3, pp. 4844-4853, June 2019.

\bibitem{intro_slam5} R. Liu, C. Yuen, T. Do, W. Guo, X. Liu and U. Tan, "Relative Positioning by Fusing Signal Strength and Range Information in a Probabilistic Framework," \textit{2016 IEEE Globecom Workshops (GC Wkshps)}, Washington, DC, 2016, pp. 1-6.
\end{comment}
\bibitem{gps} A. Yassin, Y. Nasser, M. Awad, A. Al-Dubai, R. Liu, C. Yuen, R. Raulefs, ``Recent advances in indoor localization: a survey on theoretical approaches and applications,'' in \textit{IEEE Communications Surveys \& Tutorials}, vol. 19, no. 2, pp. 1327-1346, Secondquarter 2017. 

\bibitem{intro_slam6} R. Liu, C. Yuen, {T.-N.} Do, D. Jiao, X. Liu and {U-X}. Tan, ``Cooperative relative positioning of mobile users by fusing IMU inertial and UWB ranging information,'' \textit{2017 IEEE International Conference on Robotics and Automation (ICRA)}, Singapore, 2017, pp. 5623-5629.

\bibitem{ble} N. Yu, X. Zhan, S. Zhao, Y. Wu and R. Feng, ``A precise dead reckoning algorithm based on Bluetooth and multiple sensors,'' \textit{IEEE Internet of Things Journal}, vol. 5, no. 1, pp. 336-351, Feb. 2018.

\bibitem{lte} K. Ismail, R. Liu, J. Zheng, C. Yuen, Y. L. Guan and {U-X}. Tan, ``Mobile robot localization based on low-cost LTE and odometry in GPS-denied outdoor environment,'' \textit{2019 IEEE International Conference on Robotics and Biomimetics (ROBIO)}, Dali, China, 2019, pp. 2338-2343.

\bibitem{wifi} R. Liu, C. Yuen, {T.-N.} Do, Y. Jiang, X. Liu and {U-X}. Tan, ``Indoor positioning using similarity-based sequence and dead reckoning without training,'' \textit{2017 IEEE 18th International Workshop on Signal Processing Advances in Wireless Communications (SPAWC)}, Sapporo, 2017, pp. 1-5.

\bibitem{fprint1} Y. Li, S. Williams, B. Moran and A. Kealy, ``A probabilistic indoor localization system for heterogeneous devices,'' in \textit{IEEE Sensors Journal}, vol. 19, no. 16, pp. 6822-6832, Aug. 2019.

\bibitem{fprint2} B. Jang and H. Kim, ``Indoor positioning technologies without offline fingerprinting map: a survey,'' in \textit{IEEE Communications Surveys \& Tutorials}, vol. 21, no. 1, pp. 508-525, Firstquarter 2019.

\bibitem{wifi_enhenced_mapping1} Z. S. Hashemifar, C. Adhivarahan, A. Balakrishnan, and K. Dantu, ``Augmenting visual SLAM with Wi-Fi sensing for indoor applications,'' in \textit{Autonomous Robots}, vol. 43, pp. 2245–2260, July 2019.
\bibitem{wifi_enhenced_mapping2} C. Adhivarahan and K. Dantu, ``WISDOM: WIreless Sensing-assisted Distributed Online Mapping,'', \textit{2019 International Conference on Robotics and Automation (ICRA)}, Montreal, Canada, May 20–24 2019, pp. 8026–8033.

\bibitem{propmodel} A. Zayets and E. Steinbach, ``Robust WiFi-based indoor localization using multipath component analysis,'' \textit{2017 International Conference on Indoor Positioning and Indoor Navigation (IPIN)}, Sapporo, 2017, pp. 1-8.

\bibitem{gplvm1} B. Ferris, D. Fox, and N. Lawrence, ``WiFi-SLAM using Gaussian process latent variable models,'' \textit{2007 20th International Joint Conference on Artificial Intelligence}, Hyderabad, India, Jan. 2007, pp. 2480-2485. 

\bibitem{gplvm2} J. Huang, D. Millman, M. Quigley, D. Stavens, S. Thrun and A. Aggarwal, ``Efficient, generalized indoor WiFi GraphSLAM,'' \textit{2011 IEEE International Conference on Robotics and Automation (ICRA)}, Shanghai, 2011, pp. 1038-1043.

\bibitem{ml} R. Elbasiony and W. Gomaa, ``WiFi localization for mobile robots based on random forests and GPLVM,'' \textit{2014 13th International Conference on Machine Learning and Applications}, Detroit, MI, 2014, pp. 225-230.

%\bibitem{toa} D. Frisch and U. D. Hanebeck, ``Association-free multilateration based on times of arrival,'' \textit{2020 IEEE International Conference on Robotics and Automation (ICRA)}, Paris, France, 2020, pp. 1294-1300.

%\bibitem{tdoa} M. Sun, L. Yang and D. K. C. Ho, ``Efficient joint source and sensor localization in closed-form,'' in \textit{IEEE Signal Processing Letters}, vol. 19, no. 7, pp. 399-402, July 2012.

%\bibitem{aoa} A. Ledergerber, M. Hamer and R. D'Andrea, ``Angle of arrival estimation based on channel impulse response measurements,'' \textit{2019 IEEE/RSJ International Conference on Intelligent Robots and Systems (IROS)}, Macau, China, 2019, pp. 6686-6692.

\bibitem{lidar1} H. Ye, Y. Chen and M. Liu, ``Tightly coupled 3D LiDAR inertial odometry and mapping,'' \textit{2019 International Conference on Robotics and Automation (ICRA)}, Montreal, QC, Canada, 2019, pp. 3144-3150. 
\bibitem{lidar2} G. Grisetti, C. Stachniss and W. Burgard, ``Improved techniques for grid mapping with Rao-Blackwellized particle filters,'' in \textit{IEEE Transactions on Robotic}s, vol. 23, no. 1, pp. 34-46, Feb. 2007.

\bibitem{lidar3} S. Kohlbrecher, O. von Stryk, J. Meyer and U. Klingauf, ``A flexible and scalable SLAM system with full 3D motion estimation,'' \textit{2011 IEEE International Symposium on Safety, Security, and Rescue Robotics}, Kyoto, 2011, pp. 155-160.

\bibitem{lidar4} W. Hess, D. Kohler, H. Rapp and D. Andor, ``Real-time loop closure in 2D LiDAR SLAM,'' \textit{2016 IEEE International Conference on Robotics and Automation (ICRA)}, Stockholm, 2016, pp. 1271-1278.

\bibitem{lidar5} W. {Kim}, M. S. {Ramanagopal}, C. {Barto}, M. {Yu}, K. {Rosaen}, N. {Goumas}, R. {Vasudevan} and M. {Johnson-Roberson}, ``PedX: benchmark dataset for metric 3-D pose estimation of pedestrians in complex urban intersections,'' in \textit{IEEE Robotics and Automation Letters}, vol. 4, no. 2, pp. 1940-1947, April 2019.

\bibitem{binning} R. Liu, S. H. Marakkalage, M. Padmal, T. Shaganan, C. Yuen, Y. L. Guan, {U-X}. Tan, ``Collaborative SLAM based on WiFi fingerprint similarity and motion information,'' in \textit{IEEE Internet of Things Journal}, vol. 7, no. 3, pp. 1826-1840, March 2020.
\bibitem{gmapping_vs_carographer} R. Yagfarov, M. Ivanou and I. Afanasyev, ``Map comparison of Lidar-based 2D SLAM algorithms using precise ground truth,'' \textit{2018 15th International Conference on Control, Automation, Robotics and Vision (ICARCV)}, Singapore, 2018, pp. 1979-1983.
\end{thebibliography}
\end{document}